\documentclass[letterpaper, 10 pt, conference]{ieeeconf}  

\IEEEoverridecommandlockouts                              

\newcommand{\RNum}[1]{\uppercase\expandafter{\romannumeral #1\relax}}

\usepackage{graphics} 
\usepackage{graphicx}
\graphicspath{{./figure/}}
\usepackage{amsfonts}%
\usepackage{bm}
\usepackage[noadjust]{cite}
\usepackage{hyperref}
\usepackage{amsmath} 

\usepackage{amssymb}  
\usepackage{algorithm, algpseudocode}
\usepackage{booktabs}
\usepackage{multirow}
\usepackage{cleveref}
\usepackage{color}
\usepackage{array}
\newcolumntype{C}[1]{>{\centering\arraybackslash}m{#1}}
\newcolumntype{N}{@{}m{0pt}@{}}

\usepackage{tabu}
\usepackage{siunitx}

\title{
	A Tactile Sensing Foot for Single Robot Leg Stabilization
}

\author{Guanlan Zhang$^{1}$, Yipai Du$^{2}$, ~Yazhan Zhang$^{1}$ and Michael Yu Wang$^{3}$,~\IEEEmembership{Fellow,~IEEE}%

\thanks{This work is supported, in part, by the HKSAR Innovation and Technology Fund (ITF) ITS-104-19F.} 
\thanks{$^{1}$G. Zhang, Y. Zhang (Corresponding author) are with the Department of
Mechanical and Aerospace Engineering, Hong Kong University of Science
and Technology, Hong Kong (e-mail: guanlan.zhang@connect.ust.hk; yzhangfr@connect.ust.hk.}%
\thanks{$^{2}$Y. Du is with the Department of
Electronic and Computer Engineering, Hong Kong University of Science
and Technology, Hong Kong (e-mail: yipai.du@connect.ust.hk).}%
\thanks{$^{3}$M. Y. Wang is with the Department of Mechanical and Aerospace Engineering and the Department of Electronic and Computer Engineering, Hong Kong University of Science and Technology, Hong Kong (tel.: +852-34692544; e-mail: mywang@ust.hk).}%
}

\begin{document}

\maketitle

\begin{abstract}

Tactile sensing on human feet is crucial for motion control, however, has not been explored in robotic counterparts. This work is dedicated to endowing tactile sensing to legged robot's feet and showing that a single-legged robot can be stabilized with only tactile sensing signals from its foot. We propose a robot leg with a novel vision-based tactile sensing foot system and implement a processing algorithm to extract contact information for feedback control in stabilizing tasks. A pipeline to convert images of the foot skin into high-level contact information using a deep learning framework is presented. The leg was quantitatively evaluated in a stabilization task on a tilting surface to show that the tactile foot was able to estimate both the surface tilting angle and the foot poses. Feasibility and effectiveness of the tactile system were investigated qualitatively in comparison with conventional single-legged robotic systems using inertia measurement units (IMU). Experiments demonstrate the capability of vision-based tactile sensors in assisting legged robots to maintain stability on unknown terrains and the potential for regulating more complex motions for humanoid robots.



\end{abstract}


\section{INTRODUCTION}
\label{intro}


The ability of locomotion in various environments is critical for robots. 
As robots evolving beyond conventional industrial working scenarios, the ability to navigate through terrains that are unstructured, dynamic, and even unknown becomes significant. 
For humanoid robots, a fundamental requirement in executing tasks is maintaining balance since legged robots have naturally unstable dynamics due to their inverted-pendulum-like architecture.

When the terrain is known, the robot's gait can be predetermined using model-based methods such as in \cite{mordatch2010robust} and \cite{englsberger2015three}, or learning-based methods such as in \cite{azayev2020blind}.
However, in most practical scenarios, terrains to cover are usually unknown or at least inaccurate.
Therefore, in this case, evaluation of contact condition between the robot's feet and the ground becomes a prerequisite for the control system to perform effective adaptation to the terrain. 
Various methods have been developed to map the terrain information for robot control and planning, of which perception media include camera \cite{kanoulas2017vision}, laser scanner \cite{chestnutt2009biped} and radar \cite{kauffman2013real}. However, such remote sensing techniques are limited in accuracy and indirect that cannot reflect subtle yet important changes on contact surfaces between robot feet and terrain \cite{stone2020walking}.

\begin{figure}
	\centering
	\includegraphics[scale=0.35]{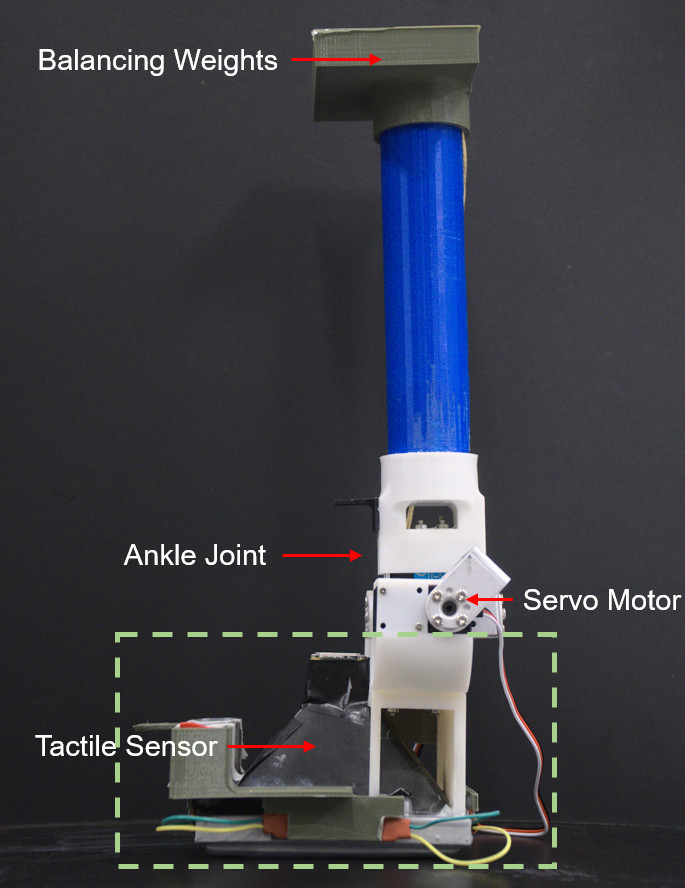}
	\caption{Prototype of the vision-based tactile sensing augmented single-legged robot.}
	\label{fig:foot}
	\vspace{-0.2cm}
\end{figure}

Inspired by the nature that haptic sensing plays an important role for animals to explore the environment, researchers apply force-torque (FT) sensors on measuring reaction force in joints and ground reaction force (GRF) on legs. The reactive force and torque signals are used for collision detection and feedback control to maintain stability against contact with obstacles \cite{kajita2010biped, morisawa2011reactive, suwanratchatamanee2009simple}. 
Apart from the proprioceptive devices used, tactile sensors for balancing have not been widely adopted in humanoids, even considering being irreplaceable in haptic systems.
As for the reasons, the underdevelopment of both hardware and algorithms are impeding the integration of tactile sensors into the humanoid stabilizing system.
Conventional tactile sensors fall short in higher-level applications, as they usually provide measurement with low resolution and inadequate dimensionality. 
Besides, interfaces of hardware and data processing become significantly complicated as the number of sensing units scales up when larger sensing areas are demanded. This also makes the fabrication process complicated and expensive. Therefore, an easy-fabricated, multidimensional, high resolution, computationally efficient tactile sensor is urgent to be developed.

Recently, vision-based tactile sensors which interface multi-axial contact information with high resolution have been emerging as a novel branch of tactile sensors \cite{chen2018tactile}. Various sensors, e.g. Gelforce \cite{sato2009finger}, FingerVision \cite{yamaguchi2016combining}, Gelsight \cite{yuan2015measurement} and Gelslim \cite{donlon2018gelslim}, are presented and applied to robotic systems in perceiving contact information.
Especially in attempting to apply tactile sensors on robot's feet, Stone et al. developed a biomimetic vision-based tactile sensor called TacTip and mounted it onto a foot of a quadrupedal robot for tactile feedback \cite{stone2020walking}. These works have demonstrated the superiority of vision-based tactile sensors in signaling important contact-related information.
In this paper, we propose the design of a vision-based tactile sensing augmented single-legged robot (shown in Fig. \ref{fig:foot}) accompanied with a contact information extraction method, which captures the interaction between ground and foot. Experiments including pose estimation of both foot and ground, and feedback control to maintain stability are conducted to demonstrate the potential of the vision-based sensor in assisting legged robot's locomotion.

This work contributes in mainly two perspectives:

\begin{itemize}
	\item Presenting a novel single robot leg design with a vision-based tactile sensing foot. 
	\item Developing a deep learning-based contact information extraction framework and presenting integration into a legged robot stabilizing system.
\end{itemize}

The paper is structured as follows. Section \RNum{2} introduces previous works related to tactile feet for legged robots and vision-based tactile sensors. In section \RNum{3}, we extensively describe the design and fabrication of the system, and the method to transform image signal to contact information. In section \RNum{4}, experiments including contact pose estimation and feedback control tasks are elaborated. We also compare our system with systems with conventional IMU sensor configurations. In section \RNum{5}, conclusions are drawn.


\section{Related Works}


\subsection{Tactile Feet for Legged Robot}

Previous attempts have been made to extend the application of tactile sensing onto legged robots and focused mainly on dynamic or static regulation with force/stress signals. Takahashi et al. \cite{takahashi2005high} designed and fabricated a high-speed pressure sensor based on thin conductive rubber to measure the distribution of normal force between the feet of a humanoid robot and the ground. Center of pressure (COP) was also measured for dynamic analysis. Other researchers explored beyond force analysis, Suwanratchatarnanee et al. \cite{suwanratchatamanee2009simple} developed a tactile sensing foot to recognize the slope of terrain by using three resistor-based transduction elements. The active adaptation was implemented to adjust the orientation of the foot and balance the robot with force signal as the feedback input. 
However, compared to the touch sensing ability of human feet, these measurements were limited in both dimensionality and resolution such that incomplete information was obtained.

With improvements on sensor technologies, high-level information can be obtained by tactile foot sensing systems. Researchers applied statistical method \cite{walas2013tactile} and learning-based methods including Support Vector Machine \cite{kolvenbach2019haptic,wu2019tactile} and Neural Networks \cite{degrave2013terrain} for terrain type classification. Guadarrama-Olvera et al. managed to detect edges of obstacle and compute supporting polygon using plantar robot skin \cite{guadarrama2018enhancing}. These works adapted algorithms of real-time complex feature extraction from measurements of basic physical contact.
In this work, we integrate a vision-based tactile sensor as a foot in a single-legged robot. Based on the system, we develop algorithms to estimate contact poses solely from the acquired tactile images that are further used as the feedback input of the feedback control to reach a balanced state.

\begin{figure}
	\centering
	\vspace{0.2cm}
    \setlength{\abovecaptionskip}{0.2cm}
	\includegraphics[scale=0.40]{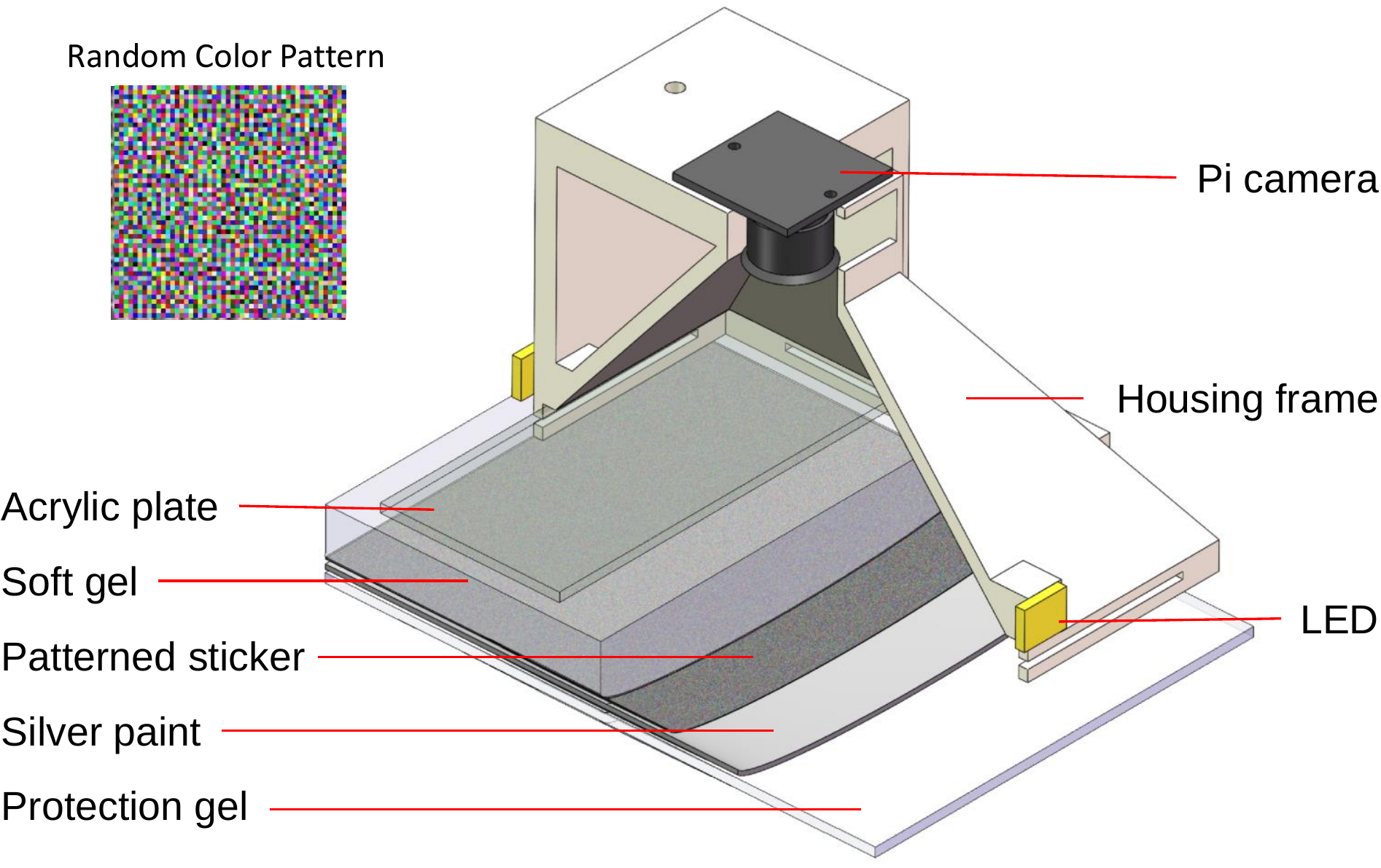}
	\caption{Schematic of sensor structure. The image on top-left is the zoomed random color pattern. Each color patch is 0.1$\times$0.1 mm$^2$.}
	\label{sensor_structure}
	\vspace{-0.3cm}
\end{figure}

\begin{figure*}
	\centering
	\includegraphics[scale=0.6]{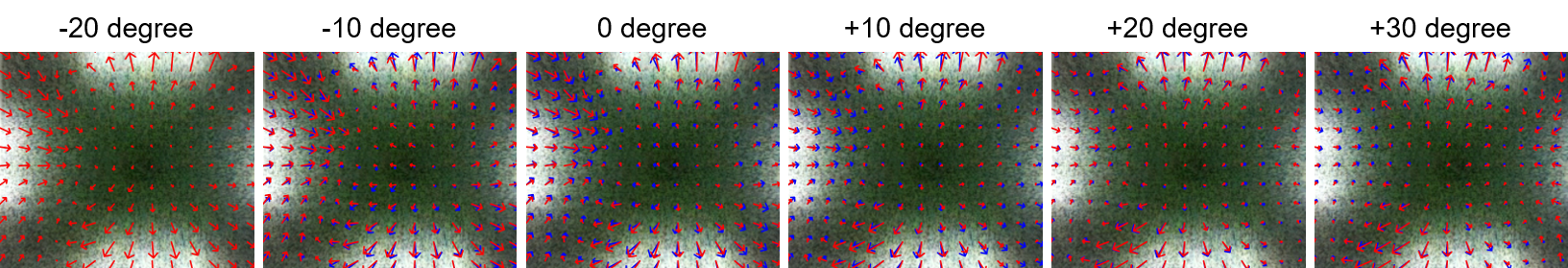}
	\caption{Visualization of DIS optical flow output. The tactile sensor is put on a ground plate with an adjustable tilting angle from -20 degrees to +30 degrees, where the plate is horizontal at 0 degrees. Six displacement vector fields corresponding to the deformation at certain degrees are shown based on the non-contact image as their reference frame. The magnitude and direction of arrows represent the displacement of the pixel. For clean visualization, we only plot vectors sparsely and scale their magnitude. Red vectors in each figure represent the optical flow in the current state, and the blue vectors represent the optical flow of the state on its left to have a better sense of the frame-to-frame difference.}
	\label{fig: opt_flow}
	\vspace{-0.3cm}
\end{figure*}


\subsection{Vision-based Tactile Sensor}

Vision-based tactile sensors are gaining more and more attention for their superior tactile resolution, easy fabrication, robust electronics, compact form-factor, and simple multiplexing peripherals.
Regarding capacity in multiple measurements and versatility in extracting various level of features, vision-based tactile sensor acquires multi-modal contact information including deformation \cite{yamaguchi2016combining}, texture \cite{johnson2009retrographic}, contact area localization \cite{donlon2018gelslim}, geometry reconstruction \cite{johnson2009retrographic} and force estimation \cite{yuan2017gelsight}\cite{sferrazza2019design}. Beyond these low-level contact information, vision-based tactile sensors have been performing effectively in high-level tasks like object recognition \cite{yuan2017connecting}, localization of dynamic object \cite{li2014localization}, simultaneous localization and mapping of the sensor on objects \cite{bauza2019tactile}, slip detection \cite{zhang2018fingervision} and a fine-grained contact events classification \cite{zhang2019towards}. 

While most previous applications of vision-based tactile sensors draw the scope of signal collection on fingertips \cite{sato2010finger}\cite{yuan2017connecting}, there exist other forms of vision-based tactile sensors. McInroe et al. presented an arm tip with integrated tactile sensing and pneumatic actuation \cite{mcinroe2018towards}. Similarly, Cramphorn et al. in \cite{cramphorn2018voronoi} fabricated a dome-shape tactile sensor called TacTip that was installed as an end effector on a manipulator. Then, Stone et al. in \cite{stone2020walking} installed TacTip on the end of the robot leg to achieve stable walking. It shows that vision-based tactile sensors are not limited to fingers and can serve other applications where high-quality contact information is required. 
Here, we investigate the capability of our previously developed vision-based tactile sensor FingerVision in estimating complete contact information for a stabilizing task of a single-legged robot to maintain balance on a tilted platform during actuation, and the results show that the sensor can be used to not just replace but even surpass conventional pose sensing configurations in humanoids.


\section{System and Methods}

In this section, a detailed description of the sensor design, fabrication process, and processing algorithms to retrieve useful contact information are elaborated. As illustrated in Fig. \ref{sensor_structure}, the tactile sensing foot is the assembly of several main components: 3D printed sensor shell, acrylic plate, elastomer skin, camera, and illumination system. The most important part is the elastomer skin, which has a multi-layer structure with a random color pattern to reflect the interaction between foot and ground in the form of displacement field signals.

The principle of the sensor is straightforward. Once the skin makes contact with external objects and deforms, the camera captures the new image of the pattern on the skin. A dense optical flow algorithm can then track the displacement of color patches on the pattern and obtain the spatiotemporal displacement field of the skin surface. Given the displacement field, features of different complexity can be extracted and utilized for various applications.
More specifications will be detailed in the following subsections.

\subsection{Tactile Sensor Design and Fabrication}
\label{sensor design}

Fabrication of the tactile foot takes account of various aspects, and we propose the design of sensor structure as shown in Fig. \ref{sensor_structure}. First, the camera's field of view (FOV) is required to cover the large area of 1000$\times$800 mm$^2$ at a distance of 80 mm. We choose Raspberry Pi Camera V2 with a fisheye lens to obtain a FOV of 160 degrees and an adjustable focal length. The camera is installed on top of the sensor housing frame. And the thickness of the housing frame is 3mm so that the strength of the sensor’s main body to withstand force from the leg is guaranteed. Then a top plate supports the ankle joint strongly with a reinforced design.  

The gel material in this sensor is a transparent, hyperelastic, and durable silicone rubber (shore hardness of 20A). Before the gel is applied, an acrylic plate is fixed inside the housing frame to support the gel. Then the solvents of two-part silicone rubber are mixed in a $1:1$ ratio and cured in the mold for casting. When formed, an elastomer layer is adhered firmly under the acrylic plate and serves as the deformation interfacing substrate. The shape of the elastomer is convex to compensate for a higher level of compression at the center region of the skin. 

In order to acquire the skin deformation, a sticker of 1050$\times$850 mm$^2$ adheres onto the surface of the elastomer layer. This random color pattern sticker, as shown in Fig. \ref{sensor_structure} top left, is printed on a thin flexible adhesive film by laserjet printers. The pattern is generated based on the principle that every color patch will have the largest distance in the RGB space compared with its 8 neighbors through random number selection in RGB space. Considering the limitation of the camera resolution and jet printing, the size of the square random color patch is 0.1$\times$0.1 mm$^2$ was achieved.
A thin frosted paint layer made of limonene, silicone, and silver paint is sprayed on the patterned sticker. It isolates potential external light disturbance, blocks background scenes, and also disperses lights from the LEDs to provide a relatively uniform illumination inside the sensor. More durable silicone rubber is coated as the protection gel to reinforce the durability of the elastomer.

Since the sensor is fully covered with opaque material, proper illumination is necessary. 8 LED lights are mounted around the elastomer layer. A semi-transparent frosted plastic plate is placed in front of each LED to generate a more diffused light. Four 3D printed parts are used to fix the LEDs. Finally, four counterweights are added in the front of the sensor housing frame to balance the center of mass.   


\subsection{Extraction of Contact Information}
\begin{figure}
	\centering
	\vspace{0.2cm}
	\includegraphics[scale=0.6]{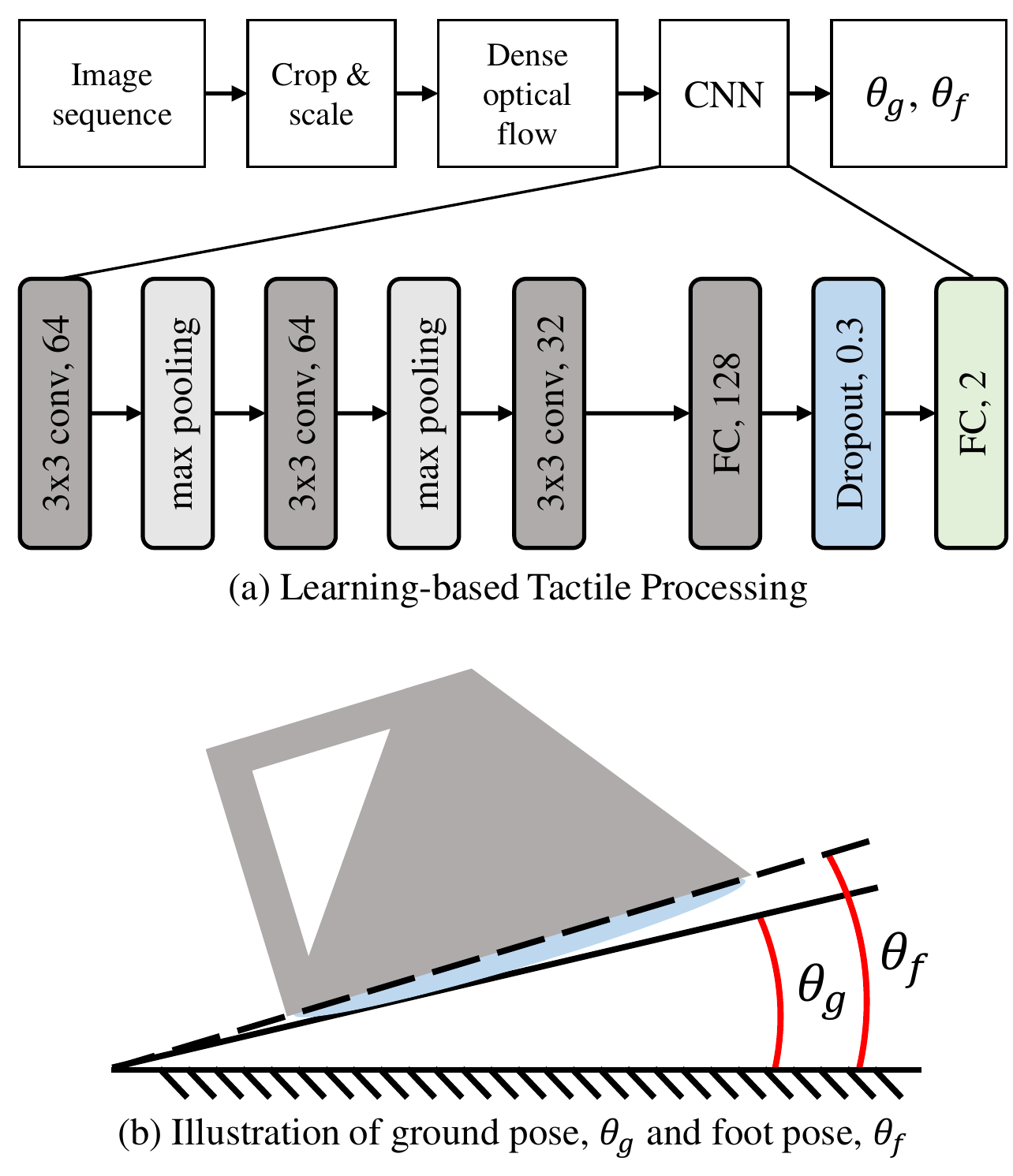}
	\caption{Pipeline of the raw image processing. (a) Learning-based tactile processing with convolutional neural network (CNN). Some abbreviations are used considering visualization. For example, "3x3 conv, 64" indicates a convolutional layer with 64 output channels and a kernel size of 3x3. And "max pooling" represents a maximum pooling layer that subsamples the input by its half size. "FC, 128" represents a fully connected layer with 128 units. "Dropout, 0.3" represents a dropout layer with 30 percent of input units dropped randomly. (b) Illustration of physical variables. $\theta_{g}$ refers to the intersection angle between the horizontal plane and inclined ground. $\theta_{f}$ refers to the intersection angle between the horizontal plane and camera frame. Two angles are usually different due to further tilting of the foot from the supporting plane.}
	\label{pipeline}
	\vspace{-0.3cm}
\end{figure}

We propose a method to extract features from the elastomer's deformation, where high-level information, such as the pose of the foot $\theta_{f}$ and slope of the ground $\theta_{g}$, are obtained for downstream applications. The pipeline of processing raw image is demonstrated in Fig. \ref{pipeline}. A sequence of image is captured by the camera with resolution of 640$\times$480 at 90 frame per second. Then they are sent to host machine with a cropped region of interest (ROI) and scaled resolution of 214$\times$182 to enhance the speed of processing in further analysis. 


\textbf{Fast Optical Flow with Dense Inverse Search (DIS)} We use the dense inverse search (DIS) optical flow algorithm \cite{kroeger2016fast} featuring more efficient computation to obtain a deformation vector field from the image sequence at a high frequency. This algorithm was first introduced to be used in vision-based tactile sensing by Sferrazza \textit{et al.} in \cite{sferrazza2019design}.
The algorithm yields optical flow \textbf{U}$_s$ by finding a warping vector \textbf{u} = (\textit{u}, \textit{v}) for each template patch \textit{T} in reference image $I_r$, which minimizes the difference between patches in reference image and query image.

\begin{equation}
\begin{aligned}
  \text{\textbf{u}} = \text{argmin}_{\text{\textbf{u}}'} \sum_{x}[I_{t}(\text{\textbf{x}} + \text{\textbf{u}}') - T(\text{\textbf{x}})]^{2}
\end{aligned}
\label{eq: warping vector}
\end{equation}
where \textbf{x} = $(x, y)^T$ is the center of patch $T$ in image $I_r$, which in our case is the initial static frame of pattern, and $I_{t}(\text{\textbf{x}} + \text{\textbf{u}}')$ is the best matched sub-window of \textit{T}(\textbf{x}) in query image $I_{t}$, which in our case, is the frame at time $t$. 

\textbf{U}$_s$ is iteratively generated from the coarsest level (with largest patch size) to the finest level (with smallest patch size). And in each iteration, the quality of optical flow is improved by Variational Refinement. 
Given the vector field, vectors with magnitudes larger than zero represent the displacement of the pattern, which is used to infer the deformation of elastomer. A brief demonstration of the output of DIS optical flow is shown in Fig. \ref{fig: opt_flow}. The tactile sensor is put on a ground plate with an adjustable tilting angle from -20 degrees to +30 degrees. The displacement vector field of pixels is computed with respect to the image without contact as a reference frame. 


\textbf{Pose estimation with neural network} The dense displacement vector fields are fed to a convolutional neural network to estimate the pose of the foot and the ground (shown in Fig. \ref{pipeline}(b)). The adoption of the deep neural network is due to the high resolution of the input displacement field, which makes model-based methods intractable. With a proper scaling factor of the tactile image, the network can be simple on structure (shown in Fig. \ref{pipeline}(a)) while sufficiently capable for the extraction purpose, which makes the training process relatively easy and quick. The dimension of the input vector field is 214$\times$182$\times$2. All convolutional layers have a stride of 1 without any padding. The activation functions are Rectified Linear Unit (ReLU).

\subsection{Single Robot Leg}

As shown in Fig. \ref{fig:foot}, we build up a single robot leg testing platform. It is used to evaluate the performance of the specially designed vision-based tactile sensor in assisting controlled stabilizing tasks to maintain a balanced state of the robot. Particularly, a $22$-centimeter-long plastic tube with a counterweight of $40$ grams on its top is connected to the tactile sensing footplate through the ankle joint. The ankle joint is then mounted on the support platform mentioned in section \ref{sensor design}, and it is driven by a servo motor to rotate at the pitching angle. The motor is controlled by a Raspberry Pi 4 Model B through a GPIO port using a PWM signal. 

In experiments, when actuated the leg is able to rotate forward/backward around the joint in $\pm 45$ degree, along which the center of mass (COM) of the entire leg-foot system changes. Changes of the COM of the whole system induce the pressure distribution and tangential dragging changes of the sensing skin if the sensor is put on a slope as illustrated in Fig. \ref{pipeline}(b). And the process deformation information of the sensor skin reflects both the slope pose $\theta_g$ and the ose $\theta_f$.


\section{Experiments and Results}

In this section, we demonstrate the working principle of the tactile sensing robot foot with a prototype shown in Fig. \ref{fig:foot} and make use of a dataset collected to train the convolutional neural network for regression. Based on the CNN model, pose estimation is obtained and integrated into closed-loop control for balancing on an inclined ground. Results of network training and feedback control are presented and analyzed.

\subsection{Data Collection and Network Training}

\begin{figure}[tbp]
	\centering
	\vspace{0.2cm}
	\includegraphics[scale=0.4]{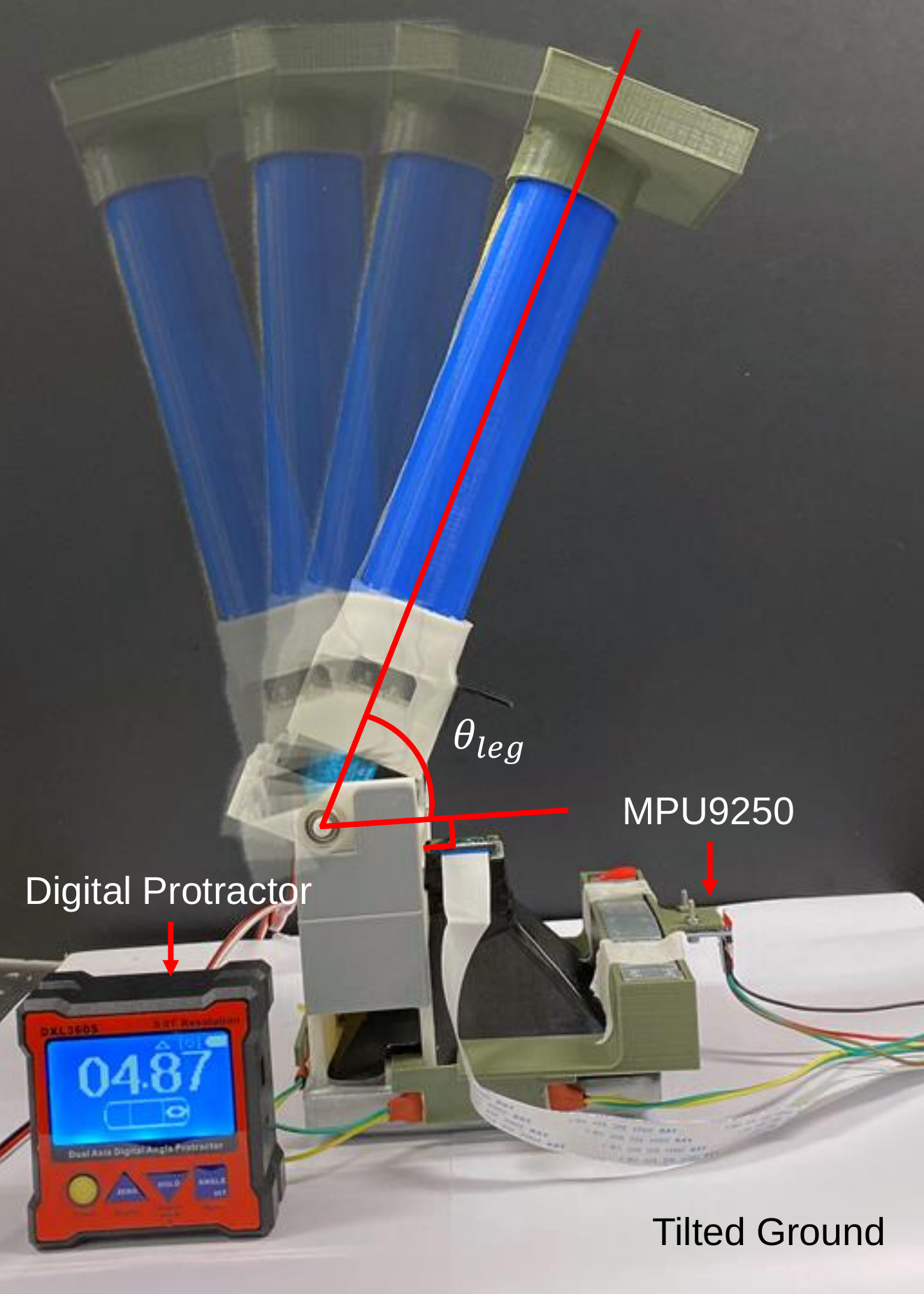}
	\caption{Setup of data collection. The tactile foot is put on a ground plate. The plate tilting angle $\theta_{g}$ is shown on digital protractor.}
	\label{fig:data_coll}
	\vspace{-0.3cm}
\end{figure}

In order to collect data for training, the tactile sensing foot with an IMU (MPU9250) mounted at its front is put on a plate (serving as the ground) with an adjustable tilting angle (shown in Fig. \ref{fig:data_coll}). Then, the following procedures are repeated.

\begin{enumerate}
\item Adjust tilting angle of the ground plate from $-12^{\circ}$ to $12^{\circ}$ with $1^{\circ}$ step size. Record the readings from a digital protractor as $\theta_{g}$. 
\item Control servo motor to rotate the leg with angle $\theta_{\text{leg}}$ from $40^{\circ}$ degree to $135^{\circ}$ with $5^{\circ}$ step size as shown in Fig. \ref{fig:data_coll}.
\item For each ($\theta_{g}$, $\theta_{\text{leg}}$) pair, save dense optical flow \textbf{U}$_{s}$ and the angle of the foot measured by MPU9250 as $\theta_{f}$. The measurement from the MPU9250 has an error in $\pm0.2$ degrees which basically meets the requirement.  
\end{enumerate}

\begin{figure}[tbp]
	\centering
	\vspace{0.5cm}
    \setlength{\abovecaptionskip}{0.2cm}
	\includegraphics[scale=0.53]{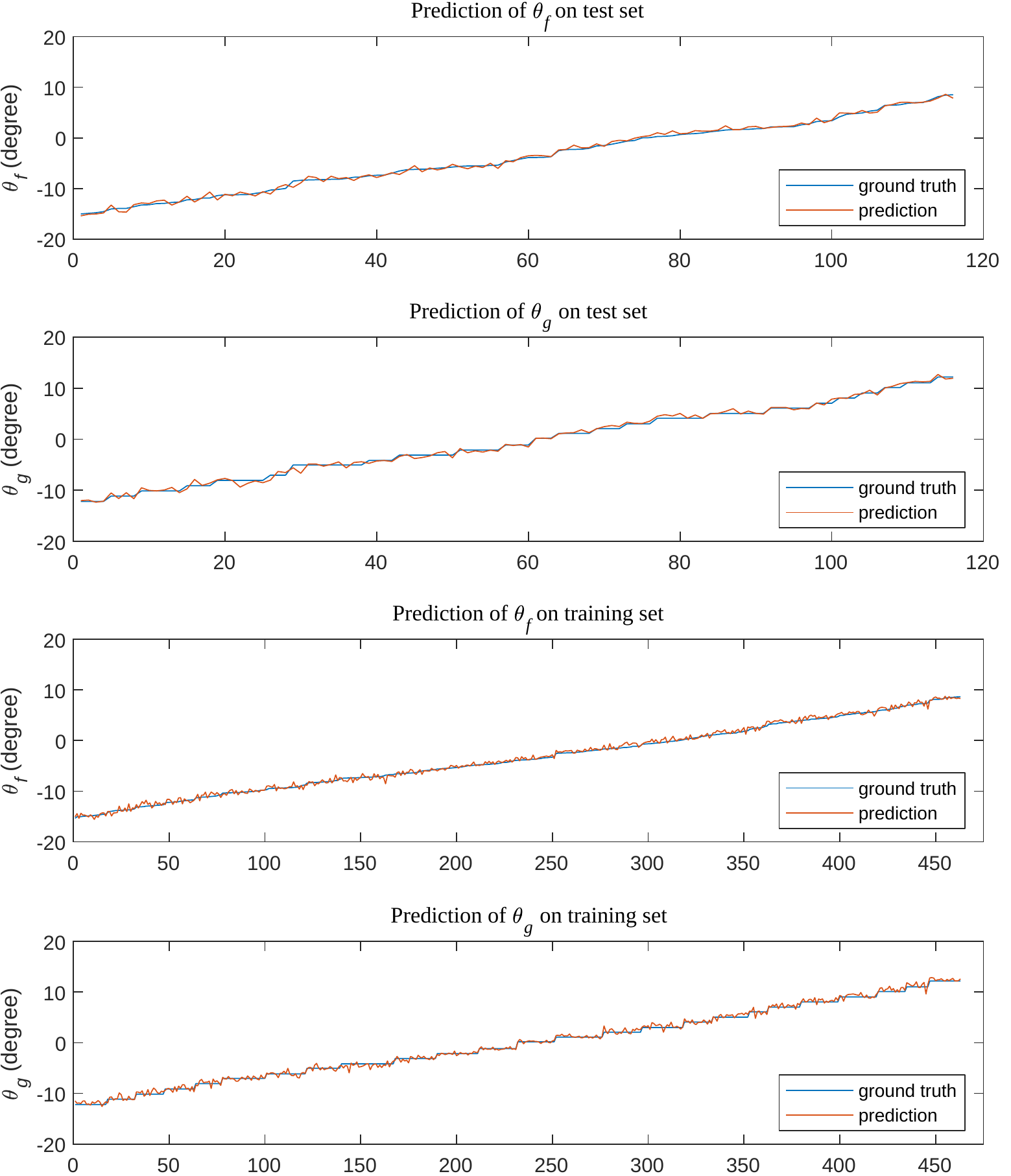}
	\caption{Prediction of $\theta_{g}$ and $\theta_{f}$ on test set and training set. The negative value indicates tilting forward and positive value indicates tilting backward.}
	\label{fig: training_result}
	\vspace{-0.3cm}
\end{figure}

We obtained $25\times20=500$ data samples, of which 80\% will be used for training and 20\% for evaluation. Adam optimizer with a learning rate of $1\times 10^{-4}$ and batch size of 16 is applied. The loss function is set to root-mean-square error (RMSE). The network was trained for $100$ epochs on a laptop with an Intel Core I7-6700HQ CPU, 16GB RAM, and an Nvidia Quadro M1000M GPU, resulting in about 45 minutes of training time using Keras framework. The Evaluations on the training set and test set are shown in Fig. \ref{fig: training_result}. We can observe that in each plot, prediction (red line) sticks tightly with ground truth (blue line), which means the sensor can predict angle in a static state with narrow error bands. 
The RMSE is shown in Table. \ref{table:1}, and it shows that our sensor can use a simple network to estimate the pose of the robot foot with low error. 

\begin{table}[htbp]
\caption{The RMSE Calculation for Training and Testing Data}
\begin{center}
\begin{tabular}{ c|c c}
\hline
\textbf{RMSE(degree)}   & $\theta_{f}$ & $\theta_{g}$ \\
\hline
Training                & 0.425        & 0.437        \\ 
Testing                 & 0.477        & 0.458        \\
\hline
\end{tabular}
\label{table:1}
\end{center}
\end{table}

\subsection{Feedback Control of Tactile Foot System}

In this experiment, we implemented real-time feedback control based on ground pose estimation to validate the proposed tactile sensor in assisting stabilization of legged robots on the ground for balancing. Specifically, two experiments were conducted including a balancing task under a quasi-static process and a comparison between systems with our tactile sensor and conventional IMU sensors.

\begin{figure}[h]
	\centering
	\vspace{0.2cm}
	\includegraphics[scale=0.26]{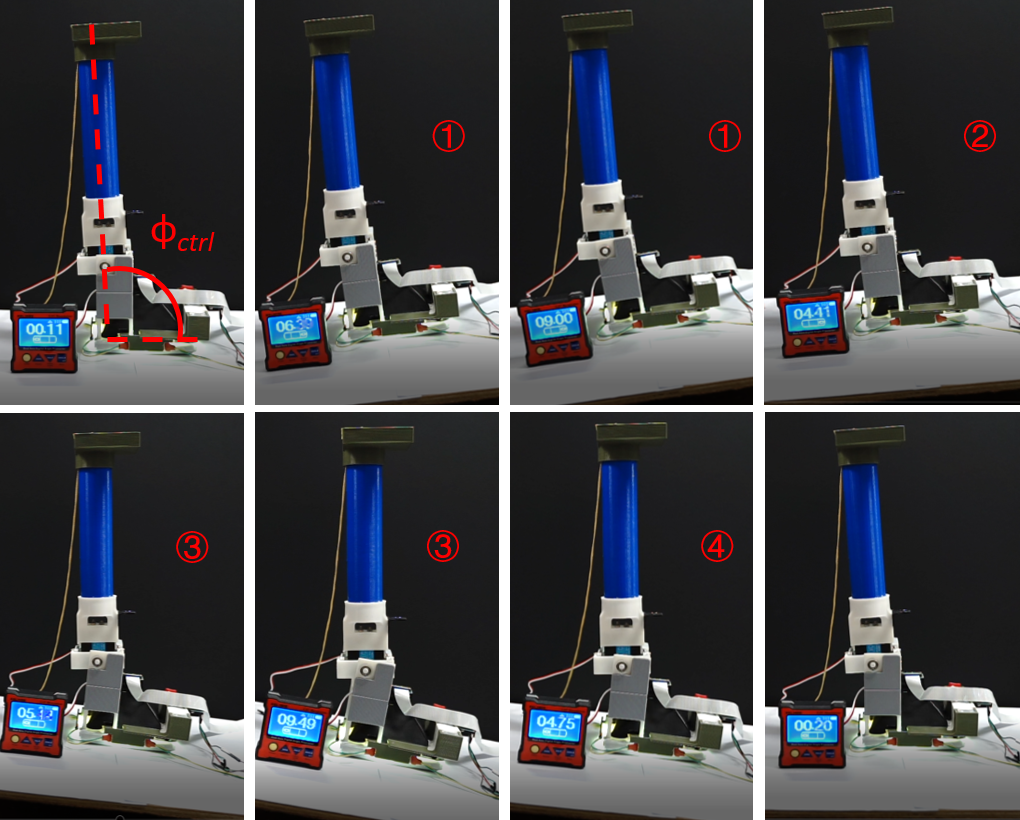}
	\caption{Demonstration of the balancing task. Angle $\phi_{ctrl}$ shown in the first image is defined as the intersection angle between the foot plate and the leg. Four stages of plate tilting operations are: 1) Tilting the ground plate to $+9$ degree; 2) Tilting the plate back to horizontal; 3) Tilting the plate to $-9.5$ degree; 4) Tilting the plate to horizontal.}
	\label{fig: real_exp}
	\vspace{-0.3cm}
\end{figure}

For the first experiment, we aim to balance the robot by rotating its leg with respect to its ankle joint as the slope of the ground changes in a quasi-static motion, during which the location of the robot's center of mass (COM) governs the system stability.
When being standstill, the robot is in balanced state thanks to the symmetric structure of the bottom part and the counterweights.
Then, to ensure the projection of COM from top view remains inside the contact area of foot and terrain, the angle between the foot plate and leg should be adjusted in reaction to the change of the ground plate angle. With feedback signals predicted by tactile sensor, $\hat{\theta}_{g}$ and $\hat{\theta}_{f}$, a desired motor angle $\phi_{ctrl}$ (shown in Fig. \ref{fig: real_exp}), can be calculated as:
\begin{equation}
\begin{aligned}
  \phi_{ctrl} = \arccos(\frac{l\cos(\hat{\theta}_{g}-\hat{\theta}_{f})}{L})+\frac{\pi}{2}-\hat{\theta}_{g},
\end{aligned}
\label{eq: angle regulation}
\end{equation}
where $L$ is the length of the leg and $l$ is the distance between the output shaft of the motor and the perpendicular bisector of the supporting platform. The motor is controlled by a 50 Hz PWM signal, with the formula of duty cycle, $D$, shown as:
\begin{equation}
\begin{aligned}
  D = K_0\times(\frac{\phi_{ctrl}}{K_1}-K_2\times\frac{d\phi_{ctrl}}{dt}+K_3),
\end{aligned}
\label{eq: PD control}
\end{equation}
where $K_0 = 0.01$, $K_1=28.8$, $K_2=0.03$ and $K_3=2.5$ are fine-tuned controller's coefficients.

\begin{figure}[tbp]
	\centering
	\vspace{0.2cm}
	\includegraphics[scale=0.2]{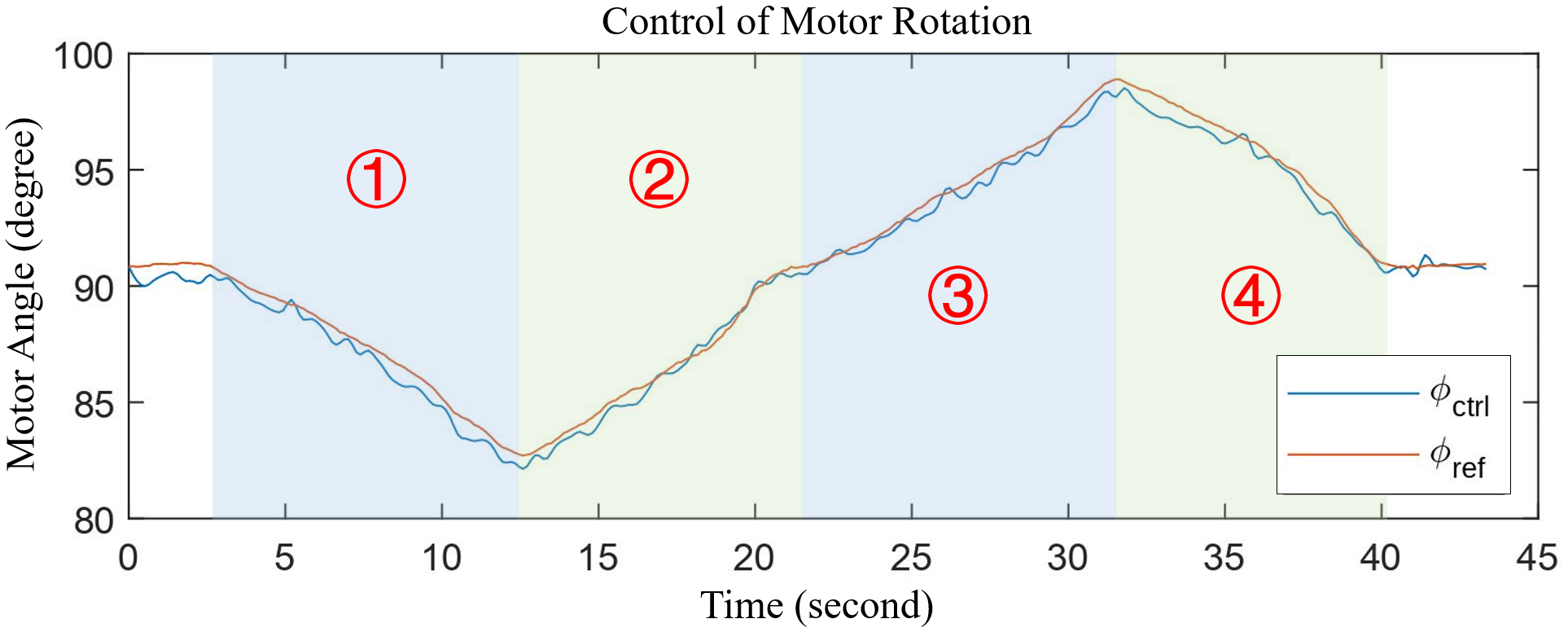}
	\caption{Result of pose regulation to minimize the error between $\phi_{ctrl}$ and $\phi_{ref}$. Four control stages of the ankle joint's angle corresponding to Fig. \ref{fig: real_exp} are marked out.}
	\label{fig: angle_regulate}
	\vspace{-0.3cm}
\end{figure}

To evaluate the system performance quantitatively, we measure the angle of foot $\theta_{f}$ from the IMU and the ground plate $\theta_{g}$ from the digital protractor to obtain $\phi_{ref}$ with Eq. \ref{eq: angle regulation}. And we want the angle of servo motor $\phi_{ctrl}$ to follow the reference angle $\phi_{ref}$, where the evaluation criteria are the error between $\phi_{ref}$ and $\phi_{ctrl}$. As shown in Fig. \ref{fig: real_exp}, four ground plate tilting operations are carried out in sequence, while the robot leg applying the control in Eq. \ref{eq: PD control} to keep the system balanced. Results of this task in Fig. \ref{fig: angle_regulate} show that $\phi_{ctrl}$ tightly follows the desired angle $\phi_{ref}$, which yields a low RMSE of 0.86 (degrees). This indicates that from a control perspective, the system response to input is satisfactory in terms of accuracy and response time. Also from the inspection of system feedback performance in each image of Fig \ref{fig: real_exp}, we can directly observe that the whole system can maintain stability during the process with little oscillation.

\begin{figure}[tbp]
	\centering
	\vspace{0.2cm}
	\includegraphics[scale=0.16]{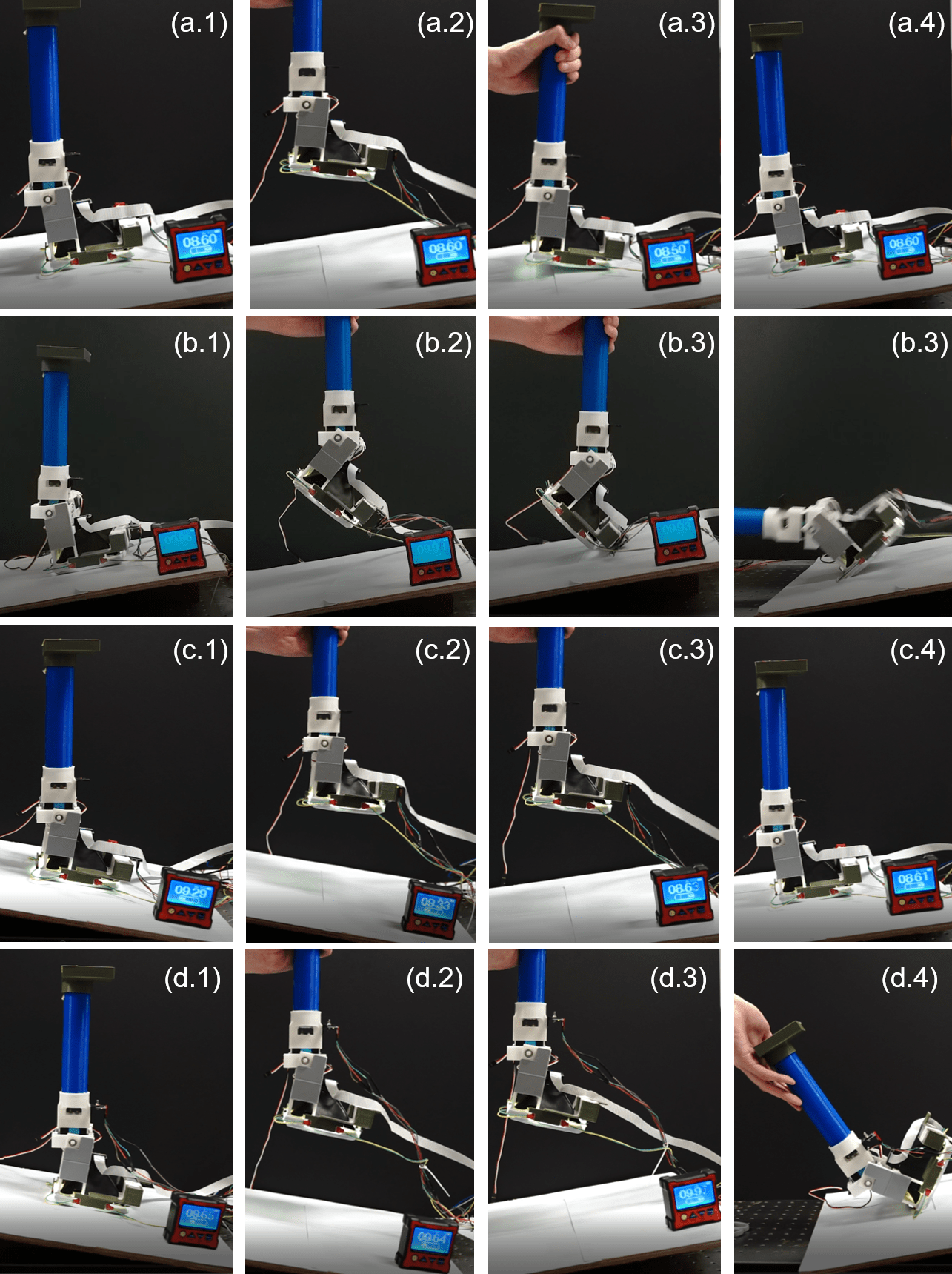}
	\caption{Comparison between system sensor configurations with tactile foot and IMU sensor in pose regulation tasks.}
	\label{fig: Compare_IMU}
	\vspace{-0.3cm}
\end{figure}

In the second experiment, we consider the scenario where a legged robot lifts and puts down its legs to complete a walking gait. Under such specific setup, we qualitatively compare the performance of a sensor with a conventional sensor in providing necessary signals for feedback control of the robot's leg (shown in Fig. \ref{fig: Compare_IMU}). Although both can measure tilting of the foot and use it as input for motor angle control, tactile sensor, being able to estimate contact information, triggers actuation only when contact occurs (Fig. \ref{fig: Compare_IMU}(a.1)). When the leg is lifted above the ground plate, the tactile skin has no deformation due to the non-contact state. Then the motor is deactivated from adjusting the leg angle (Fig. \ref{fig: Compare_IMU}(a.2)). The motor is only activated when the foot goes back onto the ground plate. With deformation detected, ground plate tilting angle can be predicted for feedback control again (Fig. \ref{fig: Compare_IMU}(a.3), (a.4)). On the other hand, the IMU sensor, as it cannot distinguish contact from the non-contact condition, will continuously try to manipulate the joint angle even when no contact is made(Fig. \ref{fig: Compare_IMU}(b.1), (b.2)). This will eventually lead to the motor rotating to a dead zone (Fig. \ref{fig: Compare_IMU}(b.3)) and failing to adapt to the original ground plate (Fig. \ref{fig: Compare_IMU}(b.4)). 

In addition, an alternative method might be to place an IMU on the leg of the foot system, so that the leg can be fixed at an absolute angle with respect to the earth's ground all the time. However, this approach fails when the foot is lifted up and then in contact with the ground plate at a different angle (Fig. \ref{fig: Compare_IMU}(d.1)-(d.3)). Because the foot remains at a tilting angle which only fits the previous ground plate condition. When it contacts the ground plate with a new angle, the system loses its stability (Fig. \ref{fig: Compare_IMU}(d.4)). But tactile sensing foot can solve this problem easily as both ground plate angle and foot angle are jointly estimated for control purpose (Fig. \ref{fig: Compare_IMU}(c.1)-(c.4)). The two cases demonstrate that tactile sensors can maintain functioning in tasks where traditional configurations with IMU sensors fail.

\section{Conclusion}


In this work, we develop a single robot leg augmented by a high-resolution vision-based tactile sensor with a novel feature extraction framework. The tactile sensor features a simple structure, easy fabrication, and reliable durability. 
An image processing pipeline is implemented to identify degrees of deformation and extract meaningful contact information. A convolutional neural network is utilized to process the tactile signal, from which estimation of ground slope and tilting of the foot can be obtained and evaluated. The sensor hardware and software integration enables tactile perception on robot feet. Evaluation experiments demonstrate the robot's capability to provide feedback control when interacting with the environment. Specifically, ground sloped estimation and active balancing control experiments are presented. Then the tactile sensing foot is compared with system configuration using only IMU sensors. Experimental results reflect the merits of the sensor designed in providing richer and more complete information solely from contact deformation of foot skin of legged robots for locomotion tasks.


\bibliographystyle{IEEEtran}
\bibliography{reference}

\end{document}